\documentclass[letterpaper, 10 pt, conference]{ieeeconf}  % Comment this line out if you need a4paper

\IEEEoverridecommandlockouts                              % This command is only needed if 
\usepackage{cite}
\usepackage{booktabs}
\usepackage{amsmath,amssymb,amsfonts}
\usepackage{graphicx}
\usepackage{textcomp}
\usepackage{duckuments}
\usepackage{xcolor}
\usepackage{soul}
\usepackage{subcaption}
\usepackage[ruled,linesnumbered, vlined]{algorithm2e}
\DontPrintSemicolon
\SetCommentSty{myCommentStyle}

\usepackage{amsfonts}
\let\labelindent\relax
\usepackage{enumitem}
\usepackage{amsmath}
\usepackage{graphicx}
\usepackage{subfig}
\usepackage[margin=0.792in]{geometry}
\usepackage{bbm}
\usepackage{algpseudocode} 
\usepackage{multirow}
\usepackage{xcolor}
\usepackage{ragged2e}
\def\BibTeX{{\rm B\kern-.05em{\sc i\kern-.025em b}\kern-.08em
    T\kern-.1667em\lower.7ex\hbox{E}\kern-.125emX}}

\title{\LARGE \bf Weakly-supervised Learning for Physics-informed Neural Motion Planning via Sparse Roadmap}

\author{Ruiqi Ni, Yuchen Liu, and Ahmed H. Qureshi% <-this % stops a space
\thanks{Ruiqi Ni, Yuchen Liu, and Ahmed H. Qureshi are with the Department of Computer Science, Purdue University, West Lafayette, IN, USA, 47907. Email {\tt\small$\{$ni117, liu3853, ahqureshi$\}@$purdue.edu}}}

\begin{document}

\let\oldtwocolumn\twocolumn
\renewcommand\twocolumn[1][]{%
    \oldtwocolumn[{#1}{
    \begin{center}
    \vspace{-5mm}
\includegraphics[width=0.97\textwidth]{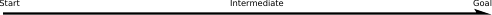}
\\[0.1cm]
%\begin{subfigure}[b]{0.97\textwidth}
\centering
\includegraphics[trim={0 0 15cm 0},clip,width=0.245\textwidth]{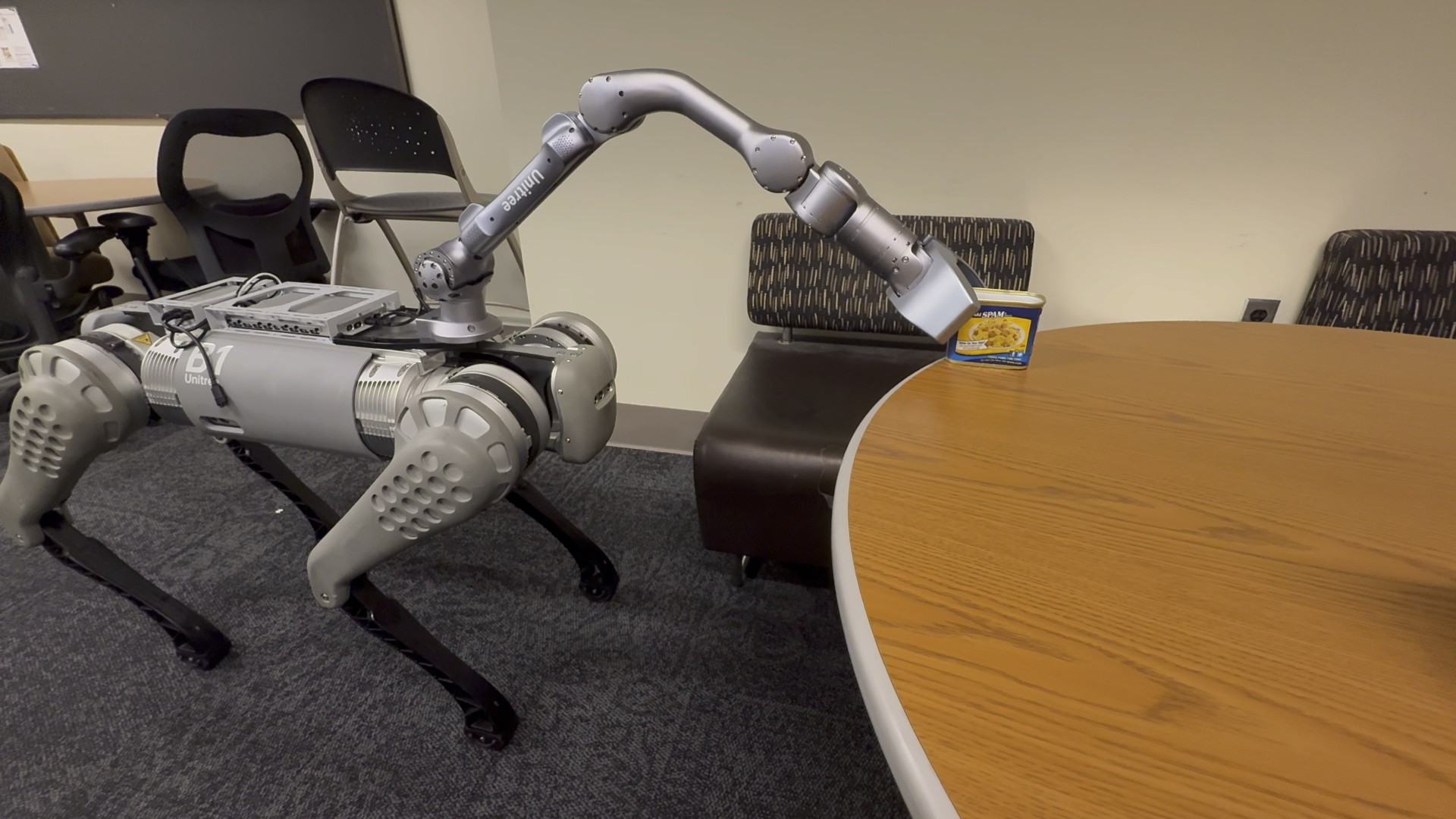}
\includegraphics[trim={6.5cm 0 8.5cm 0},clip,width=0.245\textwidth]{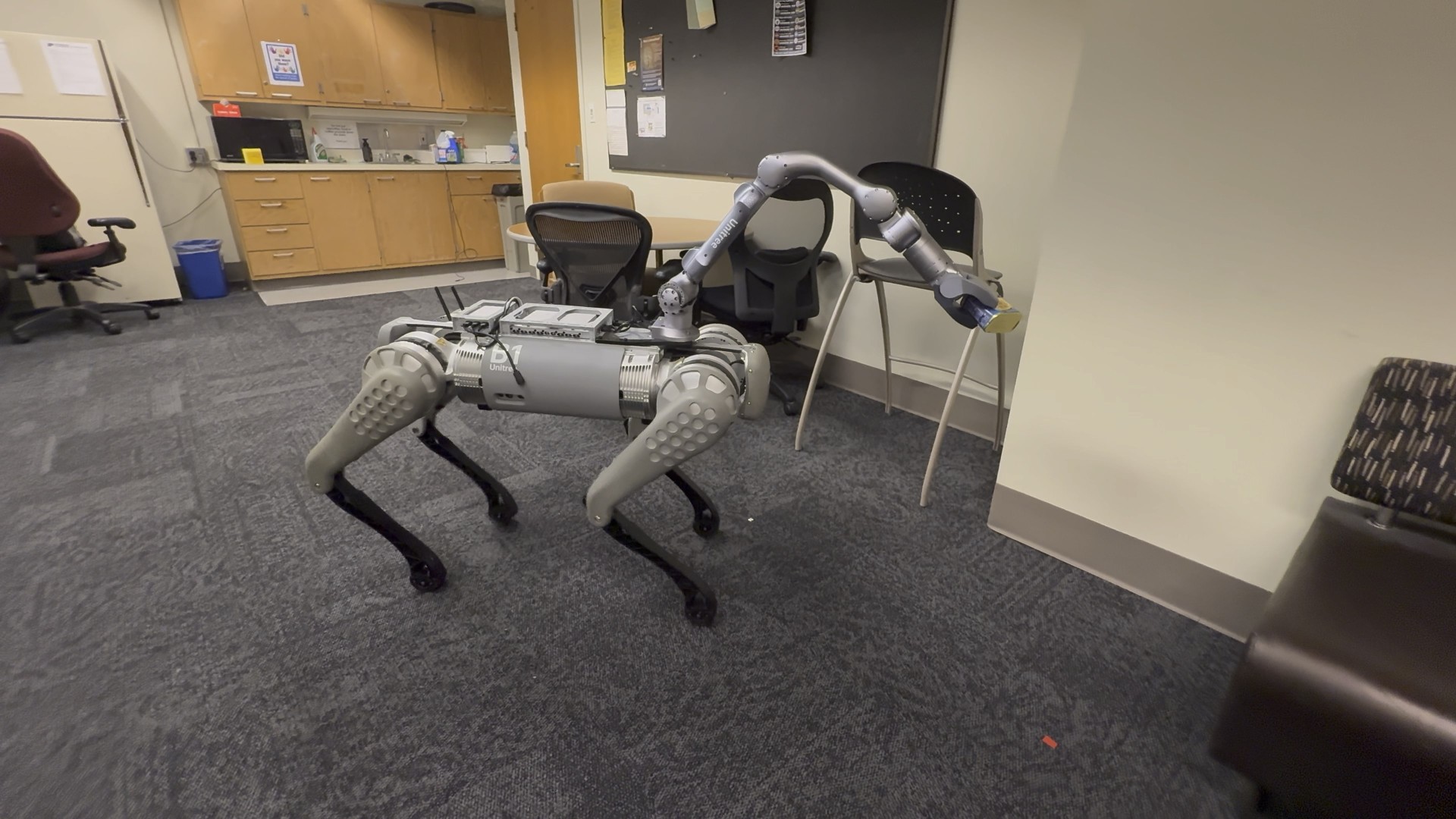}
\includegraphics[trim={12cm 0 3cm 0},clip,width=0.245\textwidth]{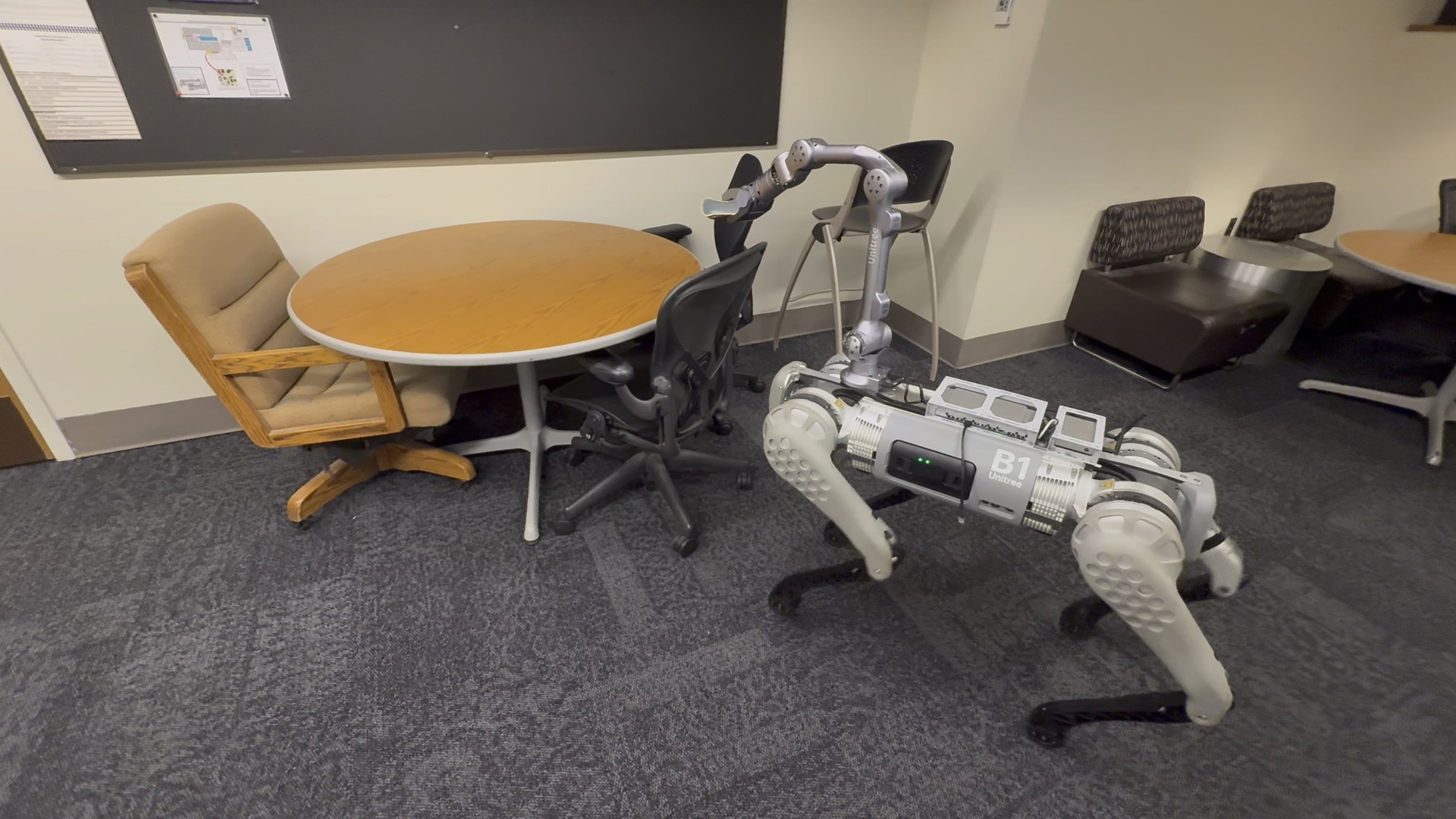}
\includegraphics[trim={10cm 0 5cm 0},clip,width=0.245\textwidth]{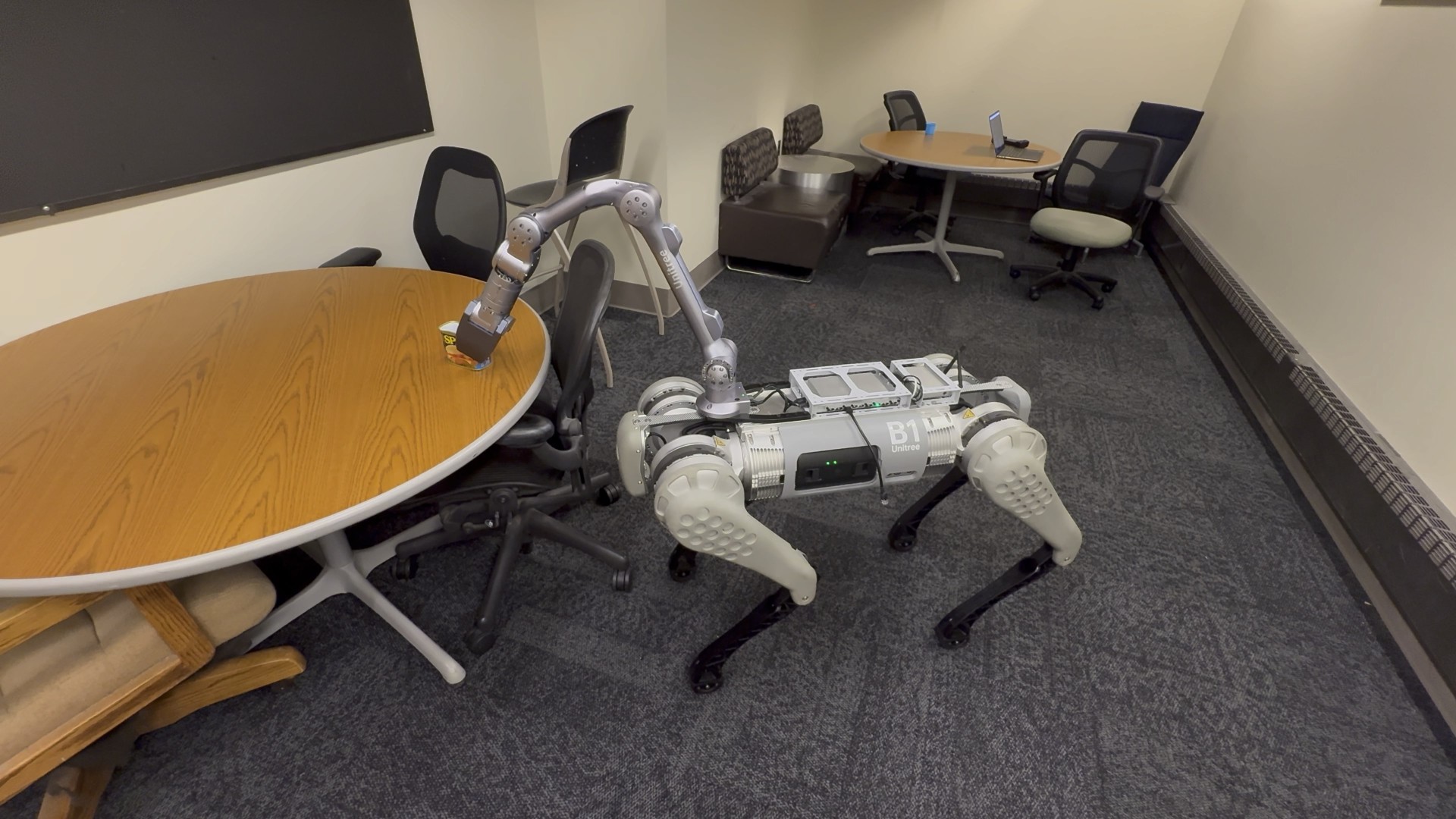}
\captionof{figure}{\justifying Demonstration of a planned path where our method enables a quadruped robot with a manipulator to transport a can between tables while avoiding collisions with a chair in a real-world kitchen environment }
\label{fig:arm_real}
\end{center}
    }]
}
\maketitle
\thispagestyle{empty}
\pagestyle{empty}

\begin{abstract}

The motion planning problem requires finding a collision-free path between start and goal configurations in high-dimensional, cluttered spaces. Recent learning-based methods offer promising solutions, with self-supervised physics-informed approaches such as Neural Time Fields (NTFields) solving the Eikonal equation to learn value functions without expert demonstrations. However, existing physics-informed methods struggle to scale in complex, multi-room environments, where simply increasing the number of samples cannot resolve local minima or guarantee global consistency. We propose Hierarchical Neural Time Fields (H-NTFields), a weakly-supervised framework that combines weak supervision from sparse roadmaps with physics-informed PDE regularization. The roadmap provides global topological anchors through upper and lower bounds on travel times, while PDE losses enforce local geometric fidelity and obstacle-aware propagation. Experiments on 18 Gibson environments and real robotic platforms show that H-NTFields substantially improves robustness over prior physics-informed methods, while enabling fast amortized inference through a continuous value representation.

\end{abstract}
\section{Introduction}
\label{sec:introduction}

Robot motion planning is a long-standing problem of finding a collision-free path between a robot's start and goal configurations in high-dimensional configuration spaces (C-spaces). %Classical approaches, such as grid-based search and sampling-based planners (e.g., PRM, RRT) \cite{lavalle1998rapidly,kuffner2000rrt,karaman2011sampling,janson2015fast,gammell2015batch}, are theoretically complete but often scale poorly in cluttered or high-DOF environments, where dense discretization or large roadmaps are required for reliability.
Classical approaches, such as grid-based search and sampling-based planners (e.g., PRM, RRT), offer strong theoretical guarantees but may require increased sampling density and collision checking effort in cluttered or high-DOF environments. To address these limitations, recent work has turned to learning-based methods that approximate cost-to-go or policy functions with neural networks, enabling efficient path inference at runtime.

Learning-based planners fall into two broad categories: supervised and self-supervised. Supervised methods~\cite{ichter2018learning,kumar2019lego,qureshi2018deeply,ichter2019robot,qureshi2019motion,qureshi2020motion,huh2021cost,li2022learning,fishman2023motion,dalal2024neuralmp} rely on expert demonstrations, often generated by classical planners, but face scalability challenges due to the high cost of producing data in complex spaces. Self-supervised methods instead exploit the rules of the planning problem itself. In particular, NTFields~\cite{ni2023ntfields}, and their extensions~\cite{ni2023progressive,ni2025physicsinformed}, approximate the solution of the Eikonal equation, where the travel-time function behaves like an optimal value function for motion planning. Physics-inspired PDE regularization enables these methods to learn without expert data while enforcing geometric fidelity and obstacle-aware propagation.

Despite their promise, existing physics-informed methods face fundamental scalability issues. In complex or multi-room environments, PDE regularization alone struggles to propagate information over long horizons, often collapsing into local minima or underestimating travel times. Supervised approaches could, in principle, provide global supervision, but they require dense roadmaps or large datasets—expensive to generate and impractical in high-dimensional spaces. Sparse roadmaps, while compact, lack the fine-grained geometric fidelity needed to enforce PDE constraints.

We propose H-NTFields, a hybrid framework that combines sparse topological supervision with physics-informed PDE regularization. PDE losses ensure smooth, obstacle-aware local propagation, while roadmap-derived supervision anchors the value function to valid long-range structures. Specifically, we build sparse roadmaps via free-space volume sampling, perturb start–goal pairs within associated free spheres to expose local obstacle geometry, and supervise the neural field with both PDE losses and weak bounds from roadmap distances. At inference, the learned value function is embedded in a sampling-based model predictive control (MPC) scheme \cite{williams2016aggressive,bharadhwaj2020model}, improving robustness and enabling integration with task-level costs.

Our contributions are as follows:
\begin{itemize}[leftmargin=*]
\item We propose a roadmap-augmented PDE training framework that unifies local geometric fidelity with sparse global structure, enabling scalable learning of time fields in complex environments.
\item We extend TD-NTFields by combining Eikonal, temporal-difference, and normal-alignment losses with roadmap-derived weak supervision, stabilized by a causality-based curriculum.
\item We integrate the learned value function into a sampling-based MPC scheme, which improves inference robustness and supports multimodal path generation.
\item We demonstrate that H-NTFields achieve substantially higher success rates than PDE-only or roadmap-only baselines, scaling to multi-room Gibson environments and high-DOF manipulators with efficient training.
\end{itemize}

H-NTFields is not intended to replace classical sampling-based planners, which provide strong theoretical guarantees. Instead, it learns a reusable continuous cost-to-go representation that can be amortized across repeated planning queries and integrated into MPC pipelines, complementing rather than competing with classical methods.

\section{Related Work}
\label{sec:related}
Research in robot motion planning spans classical search and sampling-based algorithms, trajectory optimization, and learning-based methods.

Sampling-based planners such as PRM and RRT \cite{lavalle1998rapidly,kuffner2000rrt,karaman2011sampling} approximate the connectivity of the configuration space by generating roadmaps or trees. Variants like FMT* \cite{janson2015fast} and BIT* \cite{gammell2015batch} improve efficiency by using informed heuristics and batch sampling. 
While these algorithms offer strong theoretical guarantees, their performance in high-dimensional or cluttered spaces depends heavily on sampling density and collision checking efficiency.
Trajectory optimization approaches \cite{ratliff2009chomp,kalakrishnan2011stomp,ni2021robust,Ni2021RobustMT,yin2024stein,yin2025diverse} formulate planning as a constrained optimization problem, but are prone to local minima and require good initialization.

Supervised approaches learn from expert demonstrations generated by classical planners \cite{ichter2019robot,qureshi2019motion, fishman2023motion,dalal2024neuralmp}. These methods train samplers \cite{ichter2018learning,qureshi2018deeply}, path generators \cite{qureshi2020motion,huh2021cost}, or priors for optimization-based planners. While they achieve fast inference, their performance is bottlenecked by the cost of generating large-scale labeled data. Moreover, their scalability to unseen environments is limited by the coverage of the training dataset.

Self-supervised planners avoid expert data by leveraging physics-based principles. Physics-informed planners \cite{ni2023ntfields,ni2023progressive,ni2024physics,liu2024physics,shen2024pc,li2024riemannian,muchacho2025data,liu2025physics,ren2025physics} have emerged, solving the Eikonal equation for motion planning by minimizing equation loss on offline-sampled points, providing efficient motion planning and extending downstream applications from indoor mapping navigation to constraint motion planning and manipulation. Despite their appeal, these approaches struggle in multi-room environments: even with dense sampling, they often collapse into local minima or underestimate long-range travel times without additional structural guidance.

In parallel, work on roadmap sparsification aims to represent the topology of free space with compact graphs. Sparse roadmap and related methods \cite{simeon2000visibility, dobson2013sparse, dobson2013improving, dobson2014sparse, ichnowski2019multilevel, orthey2021sparse} achieve near-optimal path quality with provably sparse structures, while hybrid approaches reduce graph density through spanner guarantees or lazy evaluation. These methods provide efficient graph representations, but are usually intended as standalone planners. Moreover, achieving reliable coverage in high-dimensional spaces often requires dense roadmaps, making construction computationally prohibitive.

Our work bridges these directions. We construct a simple sphere-packing roadmap, not as the final planner but as a weak supervisory signal that provides upper and lower bounds on travel times. Combined with PDE-based regularization, this anchors global topology while preserving obstacle-aware local propagation. Unlike dense-roadmap supervision or purely PDE-based approaches, our hybrid design scales to cluttered, multi-room environments without the computational burden of dense planning or the data cost of expert demonstrations.

\section{Background}
\label{sec:setup}

In this section, we formalize the general robot motion planning problem and then review physics-informed methods for motion planning, focusing on NTFields and their extensions, which form the basis of our approach.

\subsection{Robot Motion Planning}

We denote the robot’s workspace as $\mathcal{X} \subset \mathbb{R}^m$, where $m \in \mathbb{N}$ is the dimension of the physical environment. The robot’s configuration space (C-space) is given by $\mathcal{Q} \subset \mathbb{R}^d$, where $d$ equals the robot’s degrees of freedom (DOFs).

The workspace is partitioned into an obstacle region $\mathcal{X}_{obs} \subset \mathcal{X}$ and its complement, the obstacle-free region $\mathcal{X}_{free} = \mathcal{X} \setminus \mathcal{X}_{obs}$. Mapping through forward kinematics, the corresponding sets in C-space are $\mathcal{Q}_{obs} \subset \mathcal{Q}$ and $\mathcal{Q}_{free} = \mathcal{Q} \setminus \mathcal{Q}_{obs}$.

The motion planning problem is to find a trajectory $\xi : [0,1] \to \mathcal{Q}_{free}$ that connects a start configuration $q_s \in \mathcal{Q}_{free}$ and a goal configuration $q_g \in \mathcal{Q}_{free}$ while satisfying $\xi(0) = q_s$, $\xi(1) = q_g$. In many formulations, the planner also seeks to minimize a cost such as path length or travel time.

\subsection{Eikonal Formulation of Travel Time}

A common perspective in physics-informed methods is to treat motion planning as a wavefront propagation problem governed by the Eikonal equation. For a start state $q_s$, the solution $T(q_s,q_g)$ represents the travel time for a wavefront originating at $q_s$ to reach goal configuration $q_g$.

The Eikonal relation connects the gradient of $T$ with the local speed $S(q)$:

\begin{equation}
\small \frac{1}{S(\boldsymbol{q}_g)} = \|\nabla_{\boldsymbol{q}_g} T(\boldsymbol{q}_s, \boldsymbol{q}_g)\|,
\label{eikonal}
\end{equation}

Here, $S(q)$ reflects how fast the wavefront can expand at $q$, which is inversely related to the proximity of $q$ to obstacles. The ground-truth speed field $S^\star$ is defined based on obstacle distances:
\begin{equation}
\small S^\star(q) = \mathrm{clip}(\frac{\mathrm{d}_{obs}(q, \mathcal{X}_{obs})}{d_{max}}, \frac{d_{min}}{d_{max}},1).
\label{speed}
\end{equation}
where $d_{obs}(q,\mathcal{X}_{obs})$ is the minimum distance between the robot (at configuration $q$) and obstacles, and $d_{\min}, d_{\max}$ are user-defined thresholds. By construction, $S^\star(q)$ is smaller when near obstacles, requiring larger arrival travel time, enforcing collision avoidance.

This viewpoint highlights that the travel-time function $T(q_s,q_g)$ behaves like an optimal value function: its negative gradient indicates the direction of shortest travel, and its global structure encodes feasible connectivity.

\subsection{Neural Time Fields: NTFields and Extensions}

Building on this foundation, a series of neural planners have been proposed: NTFields \cite{ni2023ntfields} approximate $T$ with a learnable function trained by Eikonal losses. Extensions such as P-NTFields \cite{ni2023progressive} add curriculum and viscosity terms, while TD-NTFields \cite{ni2025physicsinformed} incorporate temporal-difference losses for better global consistency. Recent work also shows that ground-truth trajectories can aid training \cite{muchacho2025data}, though only in simple 2D settings.

Despite these advances, NTField variants struggle to scale in large or multi-room environments. Even with dense sampling, PDE supervision alone cannot reliably propagate information over long horizons, often leading to local minima or underestimated travel times. To address this limitation, we propose combining sparse roadmap weak supervision with PDE regularization, achieving both global consistency and local geometric fidelity.

\section{Method: Hierarchical NTFields}
\label{sec:methods}

Our approach, Hierarchical Neural Time Fields (H-NTFields), integrates sparse roadmap supervision with physics-informed PDE regularization. Roadmaps provide long-range anchors by encoding global connectivity, while PDE losses enforce local geometric fidelity and obstacle awareness. Training queries are generated by perturbing start–goal pairs near roadmap nodes, exposing local geometry, and enhancing diversity. This complementary design enables H-NTFields to capture both global structure and fine obstacle details, scaling reliably to complex multi-room environments with far fewer samples.

\subsection{Roadmap Construction for Weak Supervision}

Our framework begins with the construction of a sparse roadmap that provides structural guidance for training the PDE-based solver. Unlike classical PRMs that target dense coverage for direct path planning, our roadmap is deliberately sparse, designed to capture the essential topology of free space while leaving geometric refinement to PDE regularization.

\paragraph{Free-space volume sampling}
We generate roadmap nodes by randomly sampling collision-free points in the configuration space and associating each with a maximized free sphere \cite{yang2004sampling,shkolnik2011sample,li2024configuration}. A new node is accepted only if it lies outside the union of existing spheres, which encourages exploration of uncovered regions. This sphere-packing strategy yields a compact but diverse set of nodes that efficiently span the environment without redundancy.

\paragraph{Sparse connectivity}
Nodes are connected via straight-line collision checks, producing a graph that encodes long-range connectivity of the free space. Although sparse, this graph captures critical structural features such as inter-room passages and obstacle-induced separations. These connections provide coarse but informative supervision on feasible travel times between distant regions.

\paragraph{Start–goal perturbation}
To construct training data, roadmap nodes are selected as provisional start and goal states. Each state is then perturbed within its associated free sphere, producing random start–goal pairs. These perturbations diversify training samples in two key ways: they expose configurations near obstacles, which highlight local geometric structure, and they span both short- and long-range connections, which reveal the global topology of the environment.

\paragraph{Supervisory signals}
From this roadmap, we extract two complementary forms of supervision. First, sparse graph distances define upper and lower bounds on the travel time between nodes, anchoring the value function within feasible limits without requiring exact labels. Second, perturbed samples provide local geometric cues, including variations in speed near obstacles and orientation of boundary surfaces. These signals establish weak but informative guidance that will later be reinforced by PDE regularization (Sec.~\ref{method:pde}).

\subsection{PDE Regularization}
\label{method:pde}
To ensure that the learned time field faithfully represents the geometry of the environment and converges stably during training, we incorporate physics-inspired regularization losses originally introduced in TD-NTFields~\cite{ni2025physicsinformed}. These terms enforce consistency with the Eikonal equation at both infinitesimal and finite scales, align predictions with obstacle geometry, and stabilize training dynamics. 

Let $q_s, q_g \in \mathcal{Q}_{free}$ denote start and goal configurations in the free configuration space, and let $T(q_s,q_g)$ denote the predicted travel time between them. The function $S^\star(q)$ represents the ground-truth speed field derived from the distance to obstacles, while $S(q)$ denotes the predicted speed. The optimal policy directions $u_s^\star$ and $u_g^\star$ are obtained from the gradients of $T$ with respect to $q_s$ and $q_g$.  

\paragraph{Eikonal loss} 
The Eikonal loss $L_E$ enforces local consistency with the Eikonal equation:
\begin{equation}
\small 
L_E=(\sqrt{S^\star(q_s)/S(q_s)}-1)^2+(\sqrt{S^\star(q_g)/S(q_g)}-1)^2,
\label{eikonal_loss}
\end{equation}
ensuring that the gradient magnitude of the predicted time field matches the inverse speed field. This provides local geometric fidelity by penalizing violations of the Eikonal constraint.

\paragraph{Temporal-difference loss} 
The temporal-difference loss $L_{TD}$ captures the Bellman principle of optimality at a finite scale by enforcing value propagation over a short time step $\Delta t$:  
\begin{equation}
\begin{aligned}
\label{td}
\small L_{TD} & = \left[T(q_s,q_g) -  {\Delta t}/{S^\star(q_g)} - T(q_s, q_g + u^\star_g \Delta t)\right]^2 \\
&+ \left[T(q_s,q_g) - {\Delta t}/{S^\star(q_s)} - T(q_s + u^\star_s \Delta t, q_g )\right]^2.
\end{aligned}
\end{equation}
This complements $L_E$, which acts only as an infinitesimal tangent constraint, and prevents local overfitting by enforcing consistency across neighboring configurations.

\paragraph{Normal alignment loss} 
The loss $L_N$ aligns predicted gradients with obstacle surface normals, obtained from the gradient of the speed field near boundaries:
\begin{equation}
\begin{aligned}
\small L_{N} &=(1-S^\star(q_s))\|S^\star(q_s)\nabla_{q_s} T(q_s,q_g) + \tfrac{\nabla_{q_s}S^\star(q_s)}{\|\nabla_{q_s}S^\star(q_s)\|} \|^2 \\
&+(1-S^\star(q_g))\|S^\star(q_g)\nabla_{q_g} T(q_s,q_g) + \tfrac{\nabla_{q_g}S^\star(q_g)}{\|\nabla_{q_g}S^\star(q_g)\|} \|^2.
\end{aligned}
\label{normal}
\end{equation}
The weighting term $(1-S^\star(q))$ ensures this constraint is only enforced near obstacles.

\paragraph{Causality weight} 
The causality weight $L_C$ promotes curriculum learning by prioritizing short paths (small $T$ values) before longer ones:
\begin{equation}
\small L_C=\exp(-\lambda_CT(q_s,q_g)).
\label{cau_loss}
\end{equation}
This reflects the natural one-way propagation of value functions and stabilizes training when paths of varying lengths are included.

\paragraph{Final PDE loss} 
The combined PDE regularization objective is:
\begin{equation}
\small L_{\text{PDE}} = \big(\lambda_E L_E + \lambda_{TD} L_{TD} + \lambda_N L_N\big)\, L_C,
\label{tdloss}
\end{equation}
where $\lambda_E$, $\lambda_{TD}$, and $\lambda_N$ balance the contributions of each component. 

In summary, PDE regularization provides complementary signals: $L_E$ and $L_N$ refine obstacle-aware local propagation, while $L_{TD}$ and $L_C$ promote long-range consistency. Under our roadmap-based sampling, perturbed start–goal pairs naturally trigger these effects—bringing queries near obstacles for $L_E$ and $L_N$, and spanning varied path lengths for $L_{TD}$ and $L_C$. This synergy allows PDE regularization to reinforce roadmap supervision.

\subsection{Integration of Roadmap and PDE Supervision}
\label{sec:integration}

PDE regularization alone enforces local geometric consistency but often fails to maintain global structure in cluttered, multi-room environments. Conversely, sparse roadmaps capture long-range topology but lack fine obstacle fidelity. Our key insight is that the two are complementary: PDE losses ensure smooth, obstacle-aware propagation, while roadmap supervision anchors value estimates to valid long-range structures.

\paragraph{Roadmap-based supervision}

To construct training queries, we first perturb roadmap nodes within their associated free spheres, generating diverse start–goal pairs. This perturbation ensures that samples lie in varied regions of the free space, including areas close to obstacles and across different spatial scales.

Given such a perturbed pair $(q_s, q_g)$, we compute weak supervision bounds using the underlying roadmap. The upper bound $T_\text{ub}$ is defined as the shortest path distance along the roadmap plus the perturbation radii of $q_s$ and $q_g$, while the lower bound $T_\text{lb}$ is defined as the roadmap distance minus the perturbation radii. These bounds provide a supervision loss:
\begin{equation}
\small L_{R} = \max\!\big(0,\,T(q_s,q_g) - T_\text{ub}\big) +
\max\!\big(0,\,T_\text{lb} - T(q_s,q_g)\big),
\end{equation}
which anchors the neural time field between feasible limits without requiring exact labels.

This weak supervision constrains the value field within a feasible corridor without requiring exact path labels, preventing misestimation of travel times. Perturbations encourage queries near obstacles and across long horizons, complementing PDE regularization and providing an efficient supervisory signal for scaling neural time fields.

Beyond bounding supervision, perturbation plays a crucial role in complementing PDE regularization. Samples drawn near obstacles provide informative gradients of the speed field $S^\star$ and obstacle normals, while long-range perturbed queries encourage global propagation across distant parts of the environment. This mixture of short- and long-range pairs enables PDE terms and roadmap supervision to reinforce one another.

\paragraph{Combined training objective}
The final loss combines PDE regularization with roadmap supervision:
\begin{equation}
\small
L = \big(\lambda_E L_E + \lambda_{TD} L_{TD} + \lambda_N L_N + \lambda_R L_{R}\big)\, L_C,
\label{ourloss}
\end{equation}
where $\lambda_E,\lambda_{TD},\lambda_N,\lambda_R$ balance the contributions of each term. PDE terms guarantee smooth and obstacle-aware local propagation, while roadmap anchors provide sparse but global guidance. In addition, we find that the weak value supervision $L_R$ works particularly well when combined with the causality weighting $L_C$, since both follow the principle of “learning smaller values before larger ones.” This alignment yields a natural curriculum that stabilizes training, improves long-horizon consistency, and accelerates convergence in cluttered multi-room environments.

\paragraph{Network architecture and training details}
We adopt the TD-NTFields architecture~\cite{ni2025physicsinformed}, which encodes start and goal states into the PirateNets structure \cite{wang2024piratenets} for enhanced performance and stability to predict $T(q_s,q_g)$. The model is trained with the AdamW optimizer for 2000 epochs with a batch size of 2000. Following NTFields, each epoch samples only a mini-batch from the dataset, ensuring that training time remains nearly constant as the dataset size grows. This design makes our method suitable for large-scale data generation and training under limited computational resources.

In practice, we observe that PDE-only training tends to fail on long-horizon tasks, while roadmap-only supervision struggles to capture obstacle geometry. The integration of both resolves these limitations, yielding stable convergence and accurate value propagation in cluttered environments.

\subsection{Sampling-based MPC for Path Inference}

After training the value function $T(q_s,q_g)$, we use it as a cost-to-go function within a sampling-based model predictive control (MPC) framework \cite{williams2016aggressive,bharadhwaj2020model}. At each step, candidate actions are sampled from a Gaussian distribution, rollouts are scored by predicted travel times, and a softmax weighting biases toward lower-cost actions. A receding-horizon strategy then selects the trajectory with minimum cumulative cost.

This sampling-based MPC improves robustness over pure gradient descent by mitigating sensitivity to local minima and naturally accommodates additional terms such as collision avoidance or dynamics constraints, making it well-suited for integration into broader planning and control pipelines.

\section{Evaluation}

\begin{figure}[t]
\centering
\includegraphics[width=1.0\linewidth]{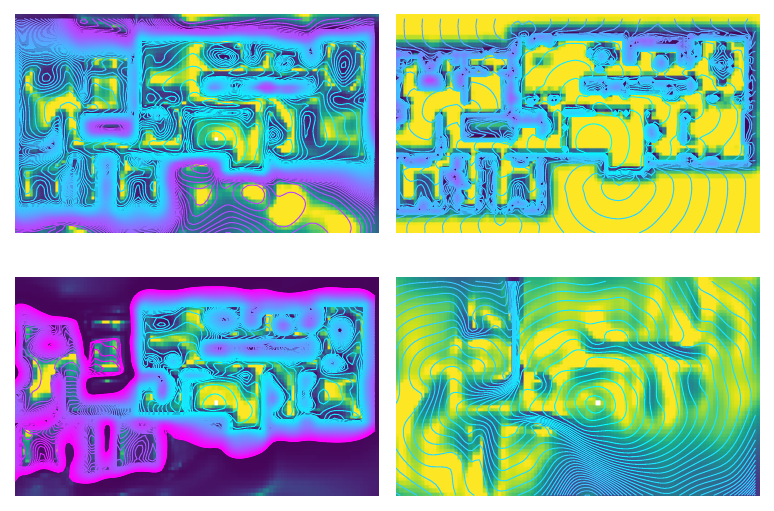}\vspace{-0.1in}
\put(-193,158){Ours}
\put(-70,158){FMM}
\put(-202,76){PDE only}
\put(-85,76){Roadmap only}
\caption{Comparison of speed and time fields in a Gibson environment. The full model produces smooth time fields with correct obstacle-aware connectivity, closely matching the ground-truth from FMM. In contrast, the PDE-only variant collapses into local minima (lower-left region), while the roadmap-only variant preserves global contours but fails to capture fine obstacle geometry.}
\label{fig:ablation-vis}\vspace{-0.2in}
\end{figure}
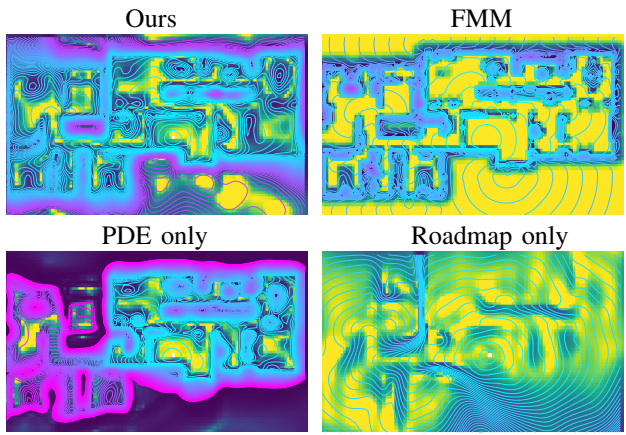

In this section, we present the experimental evaluation of our proposed H-NTFields. We begin with ablation studies to isolate the contribution of each module within our framework and to examine performance as the amount of training data increases. We then compare our method against a range of planning algorithms across 18 challenging 3D environments from the Gibson dataset \cite{li2021igibson}, highlighting its robust performance in complex settings. Finally, we demonstrate the scalability of H-NTFields on a 6-DOF manipulator and a 10-DoF mobile manipulator, showcasing its ability to solve long-horizon planning tasks in real-world scenarios. All experiments were conducted on a workstation equipped with an RTX 3090 GPU, an Intel Core i9 processor (3.50GHz × 8), and 32 GB RAM.

\begin{table}[h]\vspace{0.1in}
\centering
\scalebox{1}{ %1.2
\begin{tabular}{cccc}
\toprule
\multirow{2}{*}{Methods} & \multicolumn{3}{c}{Performance Metrics} \\ 
\cmidrule{2-4}
& \multicolumn{1}{c}{Time (sec) $\downarrow$} & \multicolumn{1}{c}{Length } & \multicolumn{1}{c}{SR (\%) $\uparrow$} \\ 
\midrule
Ours  & $0.115 \pm 0.071$ & $0.72 \pm 0.46$ & 90.8 \\
Roadmap only& $0.105 \pm 0.071$ & $0.77 \pm 0.52$ & 68.5 \\
PDE only& $0.089 \pm 0.053$ & $0.68 \pm 0.46$ & 60.0 \\
\bottomrule
\end{tabular}}
\caption{Quantitative ablation study on Gibson environments. The full H-NTFields model achieves the highest success rate while maintaining comparable planning time to the ablated variants. PDE-only reduces average path length slightly but fails in long-horizon tasks, while roadmap-only improves global consistency yet misses fine obstacle fidelity, leading to lower overall performance.}
\label{table:abl_exp}
\vspace{-0.2in}\end{table}

\subsection{Ablation Studies}

The primary objective of our ablation studies is to assess the complementary contributions of the PDE loss and the Roadmap-based weak supervision. We consider three variants: (i) PDE-only, which uses the Eikonal PDE loss without roadmap supervision in Eq. \ref{tdloss}; (ii) Roadmap-only, which relies solely on roadmap-derived upper and lower bounds without PDE constraints; and (iii) Full (Ours), which integrates both components in Eq. \ref{tdloss}. All models are trained with the same dataset of 50k start–goal pairs. To avoid this confounding factor, we adopt the NTFields’ distance-uniform sampling strategy, which allows us to isolate the effect of supervision modules under fair and consistent conditions.

We evaluate the three variants in terms of success rate (SR), path length (Length), and planning time (Time), and the metric is evaluated using 100 randomly selected
start-goal configuration pairs in free space across 18 challenging 3D environments from the Gibson dataset \cite{li2021igibson}. 

Table~\ref{table:abl_exp} reports the quantitative results of the ablation study. The full model achieves the highest success rate (90.8\%) while maintaining comparable planning time and path length to the ablated variants. PDE-only achieves slightly shorter average path length, but its overall success rate drops sharply to 60.0\%. Roadmap-only performs better (68.5\%) by leveraging global connectivity, but still lags far behind the full model.

To further illustrate these differences, Fig.~\ref{fig:ablation-vis} visualizes the predicted time fields (shown as contour lines) and speed fields (shown as color maps) in a representative Gibson environment. The contour lines are centered on the goal location (middle of the scene), and their gradient descent direction indicates the expected path toward the goal. The speed field, represented by the color map, reflects the indoor layout: high-speed regions (yellow) correspond to free space, while low-speed regions (dark bands) align with obstacle boundaries.

Our method captures both global connectivity and fine obstacle geometry: contour lines remain consistent across rooms, while the speed field aligns with free space and obstacle boundaries. PDE-only, in contrast, captures local speed variations but collapses into local minima in complex layouts, breaking long-horizon connectivity. Roadmap-only recovers a coarse approximation of the global field by leveraging long-range supervision, but without PDE regularization, it fails to reconstruct the speed field or capture obstacle boundaries in detail.

Overall, these results confirm that PDE loss and Roadmap supervision provide complementary benefits: PDE regularization ensures smooth and obstacle-aware value fields, while the roadmap supplies sparse but crucial long-range connectivity. Their integration is essential for robust performance in complex environments.

\subsection{Scalability with Supervision Size}

We next investigate the sample efficiency of H-NTFields by varying the number of supervised start–goal pairs $M \in {10k,20k,50k,100k,200k}$ across 18 Gibson environments, while fixing the roadmap size at 5,000 nodes. This setup isolates the effect of dataset size from roadmap coverage and directly tests how much supervision is needed for reliable learning.

As shown in Table~\ref{table:scale_exp}, H-NTFields already achieves 61\% success with only 10k samples—surpassing PDE-only trained with 50k—and quickly scales to 90\% at 50k. Performance then saturates, with no notable gains at 100k or 200k, showing that strong results can be reached with far fewer samples than prior work. By contrast, PDE-only requires at least 100k samples to approach 75\% success, while TD-NTFields does not exceed 77\% even with 200k.

The roadmap overhead is modest: construction and query take only a few seconds per scene, while perturbation-based sampling and PDE supervision are identical to PDE-only baselines. Training time per epoch remains nearly constant at ~2.7 minutes, since batching follows NTFields; smaller datasets further reduce memory usage, making the approach practical for edge devices.

Overall, these results highlight the sample efficiency of H-NTFields: reliable performance is achieved with ~50k pairs—roughly a quarter of what prior PDE-only methods typically require—while also delivering substantially higher final success rates and robustness in multi-room environments.

\begin{table}[h]\vspace{0.1in}
\centering
\scalebox{1}{ %1.2
\begin{tabular}{ccccc}
\toprule
\multirow{2}{*}{Data} & \multicolumn{4}{c}{Performance Metrics} \\ 
\cmidrule{2-5}
& \multicolumn{1}{c}{Sample $\downarrow$} & \multicolumn{1}{c}{Roadmap $\downarrow$} & \multicolumn{1}{c}{SR (Ours) $\uparrow$} & \multicolumn{1}{c}{SR (PDE) $\uparrow$}\\ 
\midrule
10k  & 0.23 &  14.3 & 61.4 & 20.0\\
20k  & 0.26 &  16.3 & 82.5 & 26.6\\
50k & 0.32 &  17.6 & 90.8 &59.8\\
100k & 0.42 &  18.1 & 92.7 &75.1\\ 
200k & 0.63 &  18.4 & 92.0 &76.8\\ 
\bottomrule
\end{tabular}}
\caption{Scalability study on Gibson environments. H-NTFields achieves 90.8\% success with only 50k training pairs, while TD-NTFields requires 200k samples to reach just 76.8\%—lower than H-NTFields trained with only 20k. Reported costs (seconds) include sample generation and roadmap construction/query, showing that combining sparse roadmap supervision with PDE regularization yields both efficiency and scalability.}
\label{table:scale_exp}
\vspace{-0.1in}\end{table}

\begin{figure}[t]
\centering
\begin{subfigure}{0.45\textwidth}
\centering
\includegraphics[trim={4cm 3cm 2cm 1cm}, clip,width=1.0\linewidth]{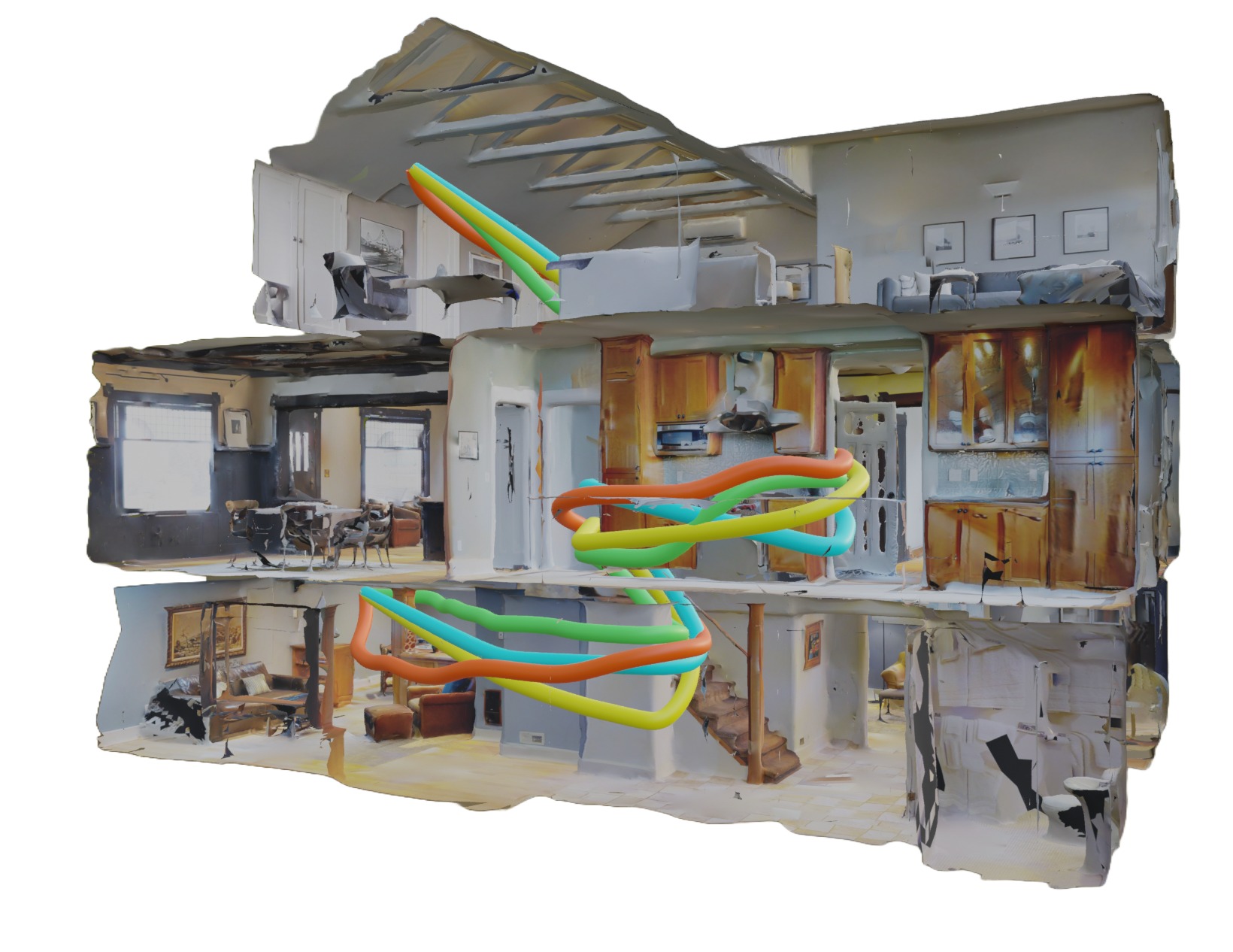}
\end{subfigure}
\caption{Path comparison in a Gibson environment. The figure shows paths generated by different planners for the same start–goal query. NTFields, P-NTFields, and TD-NTFields fail to produce viable paths. Our method successfully generates a smooth trajectory (orange) in 0.22s. RRTConnect (yellow) and Lazy-PRM (cyan) find valid paths but are slower (3.33s and 3.99s). FMM (green) runs faster (0.98s) but produces paths with some sharp turns due to grid discretization.}
\label{gibson}\vspace{-0.2in}
\end{figure}

\begin{table}[h]\vspace{0.1in}
\centering
\scalebox{1}{ %1.2
\begin{tabular}{cccc}
\toprule
\multirow{2}{*}{Methods} & \multicolumn{3}{c}{Performance Metrics} \\ 
\cmidrule{2-4}
& \multicolumn{1}{c}{Time (sec) $\downarrow$} & \multicolumn{1}{c}{Length } & \multicolumn{1}{c}{SR (\%) $\uparrow$} \\ 
\midrule
Ours  & $0.115 \pm 0.071$ & $0.72 \pm 0.46$ & 90.8 \\
NTFields  & $0.076 \pm 0.053$ & $0.75 \pm 0.94$ & 27.2 \\
P-NTFields & $0.076 \pm 0.064$ & $0.74 \pm 0.85$ & 37.2 \\
TD-NTFields & $0.081 \pm 0.051$ & $0.72 \pm 0.46$ & 76.8 \\
RRTConnect & $1.165 \pm 1.876$ & $0.71 \pm 0.41$ & 98.8 \\
LazyPRM & $0.496\pm 1.343$ & $0.68 \pm 0.38$ & 99.7 \\
FMM & $0.822 \pm 0.028$ & $0.70 \pm 0.46$ &  99.0 \\ 
\bottomrule
\end{tabular}}
\caption{Comparison for all motion planning methods in 18 Gibson environments. It can be seen that our method exhibits much higher SR than other physics-informed methods with relatively slower planning times. The relatively slow planning times are due to our method solving much harder cases, where the start and goal span across multiple rooms. }
\label{table:main_exp}
\vspace{-0.2in}\end{table}

\begin{figure*}[th]
\centering
\includegraphics[width=0.97\textwidth]{figures/arr1.pdf}
\\[0.1cm]
\begin{subfigure}[b]{0.99\textwidth}
\centering
\includegraphics[trim={9cm 0 1cm 0},clip,width=0.245\textwidth]{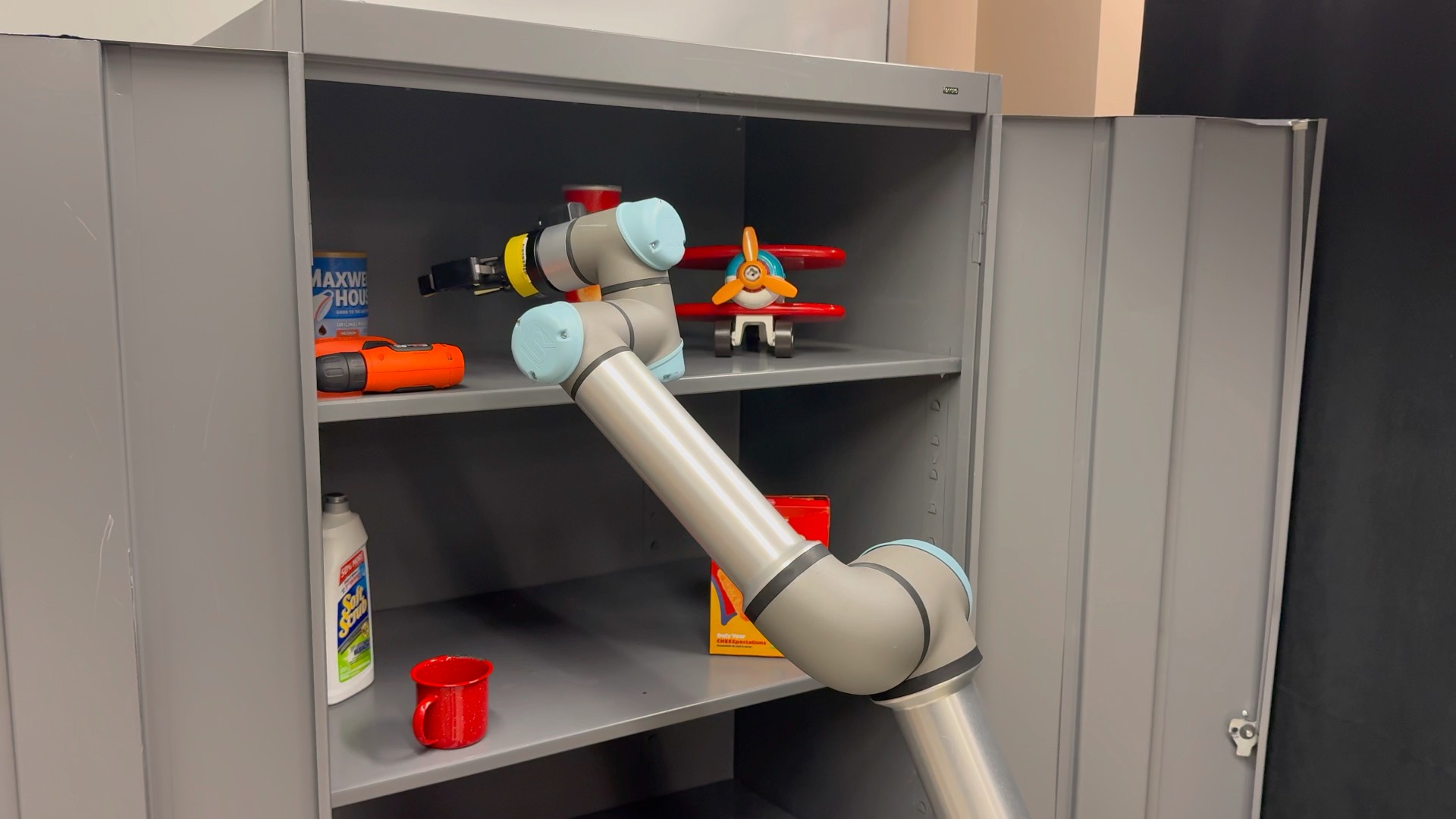}
\includegraphics[trim={9cm 0 1cm 0},clip,width=0.245\textwidth]{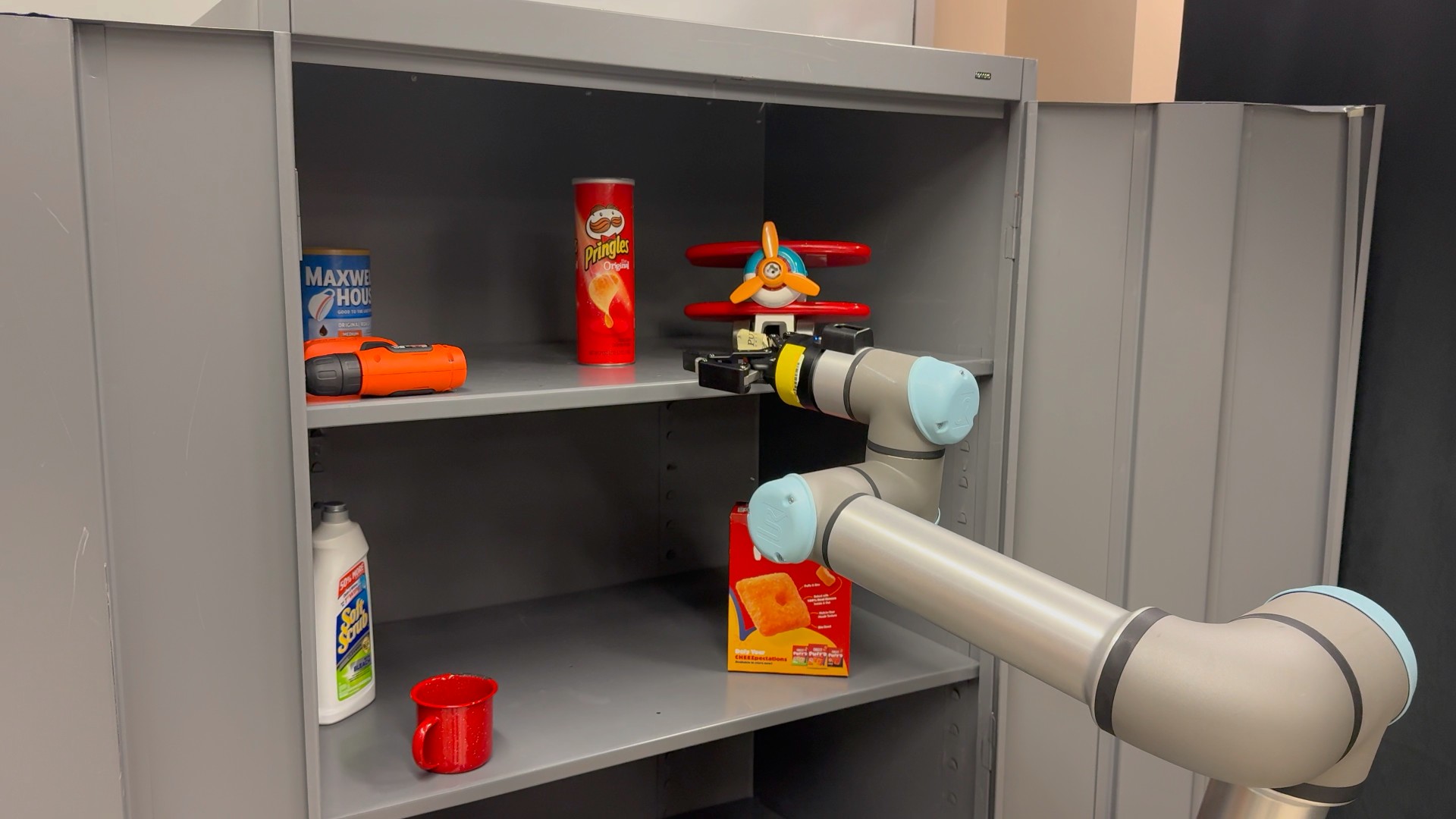}
\includegraphics[trim={9cm 0 1cm 0},clip,width=0.245\textwidth]{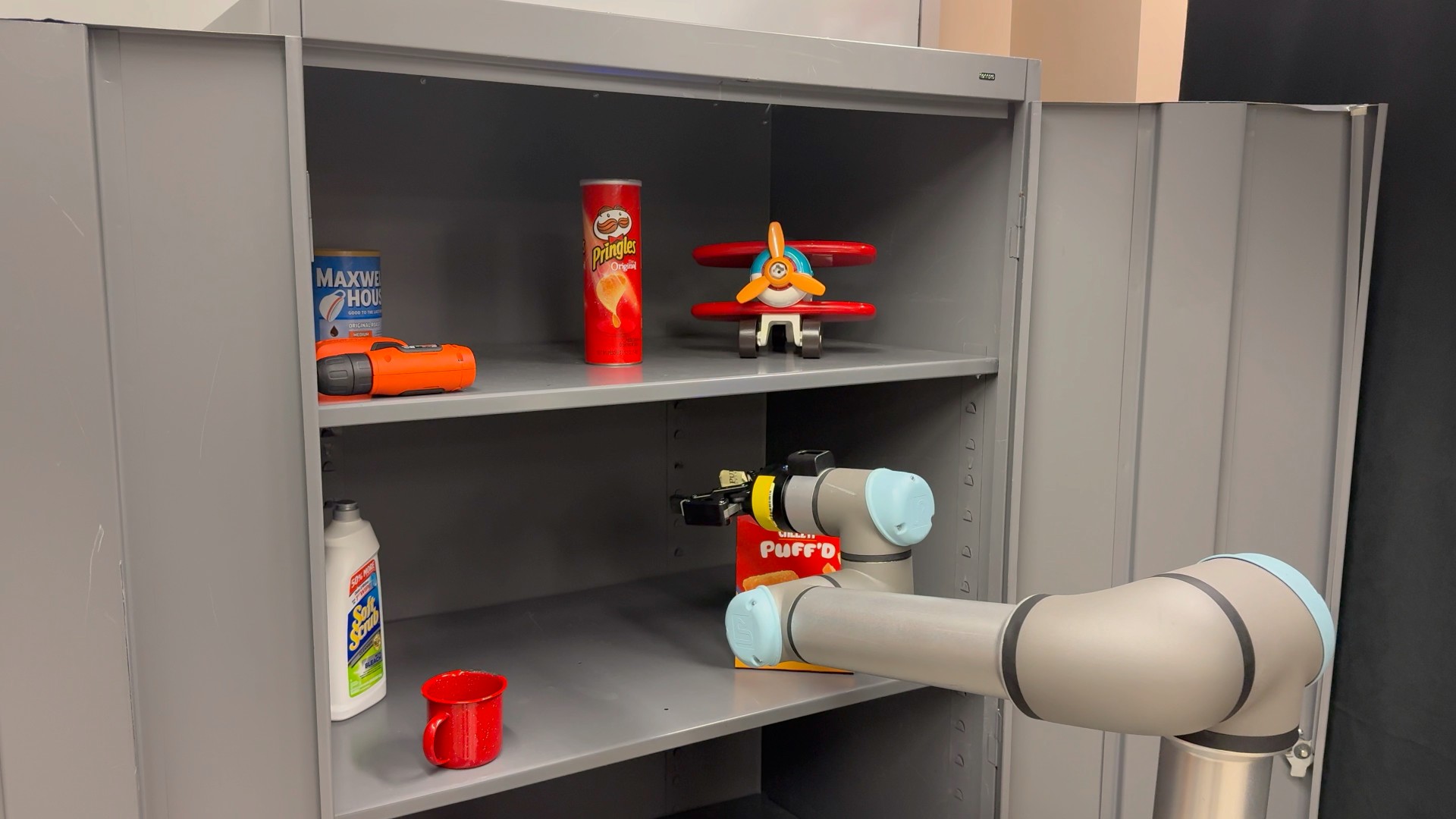}
\includegraphics[trim={9cm 0 1cm 0},clip,width=0.245\textwidth]{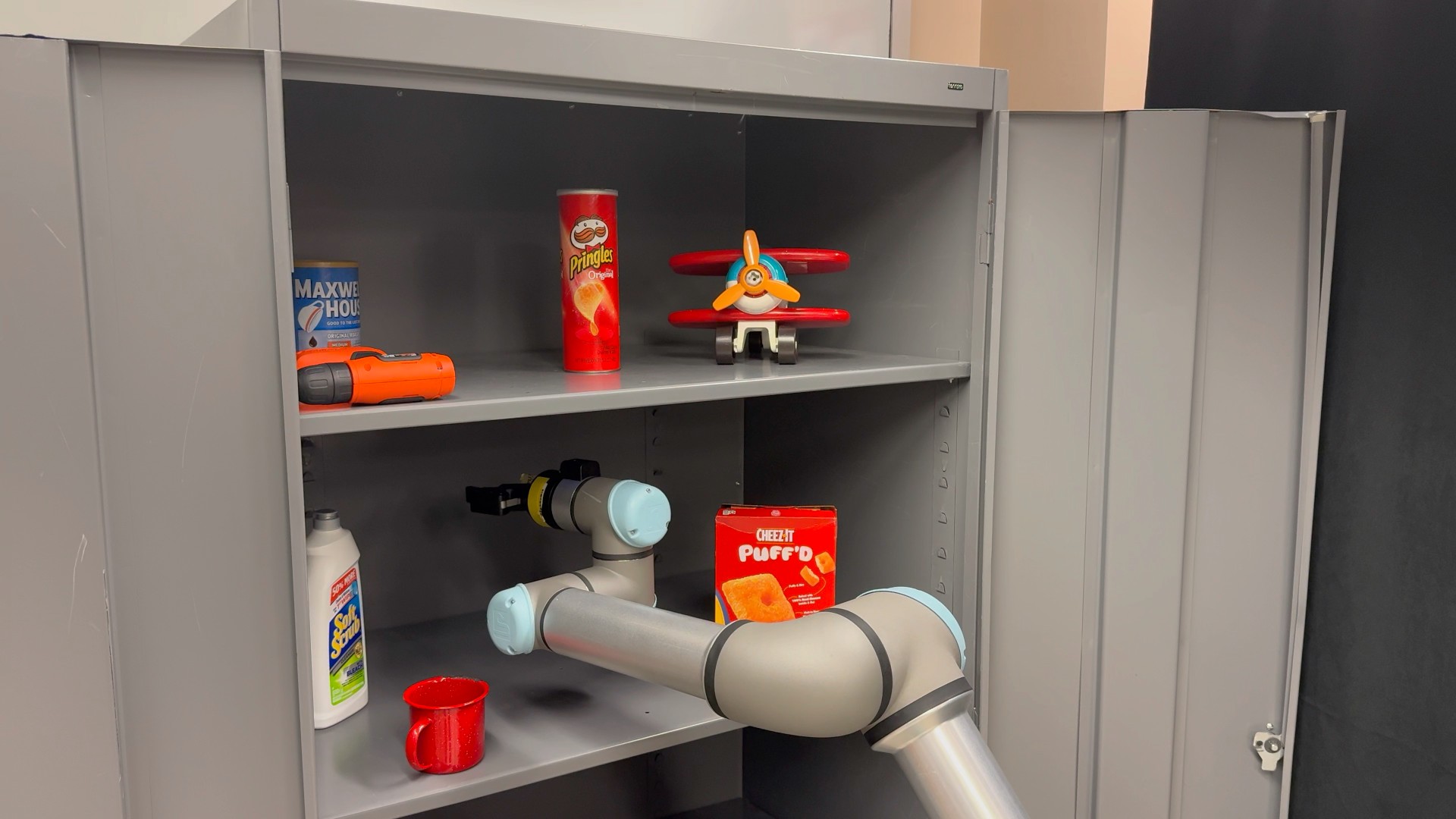}
\end{subfigure}
\captionof{figure}{\justifying Demonstration of a planned path where the robot arm navigates through a complex cabinet environment.}
\label{fig:ur5e_arm}\vspace{-0.2in}
\end{figure*}

\subsection{Comparison Analysis over Simulated Environments}

\textbf{Baselines.} We compare our method against physics-informed planners, classical sampling-based planners, and a grid-based method. NTFields \cite{ni2023ntfields}, P-NTFields \cite{ni2023progressive}, TD-NTFields \cite{ni2025physicsinformed} approaches represent prior attempts to learn continuous value functions without expert supervision by solving the Eikonal equation. Sampling-based planners (RRTConnect \cite{kuffner2000rrt} and LazyPRM \cite{bohlin2000path}) are widely used classical algorithms. The fast marching method (FMM \cite{sethian1996fast}) also solves the Eikonal PDE but relies on discretization and graph search rather than continuous learning. All methods use a KDTree-based collision checker. Neural-field methods were evaluated on GPU, while classical planners were executed on CPU using standard implementations.

\textbf{Metrics.} The evaluation metrics include planning time, path length, and success rate (SR). A time limit of 10 seconds is imposed for sampling-based methods; cases exceeding this limit are treated as failures. Planning time and path length are only reported for successful cases.

\textbf{Gibson Experiments.} We evaluate all methods across 18 Gibson environments, which feature cluttered 3D indoor layouts with long corridors, narrow passages, and up to 19 rooms spanning approximately 700 $m^2$. Each environment contains three floors, ensuring multi-level connectivity and long-horizon planning challenges. For evaluation, we sample 100 random start–goal pairs per environment within the free space, yielding a total of 1,800 queries.

Physics-informed methods offer an attractive representation for motion planning: they produce continuous value fields that enable near-instant inference and can be parallelized across thousands of queries on modern GPUs. These properties make them effective not only as planners but also as scalable data generators for downstream learning tasks.

However, as shown in Table~\ref{table:main_exp}, existing physics-informed approaches such as NTFields and P-NTFields achieve low success rates in complex Gibson environments (27.2\% and 37.2\%, respectively), despite their fast inference. TD-NTFields improves performance to 76.8\% but still fails in many multi-room settings. In contrast, H-NTFields attains substantially higher success rates with only 50k training samples, compared to the 200k typically required by prior methods. This underscores the sample efficiency gained by combining roadmap anchors with PDE regularization.

By integrating sparse roadmap supervision into PDE learning, we retain the strengths of neural fields—fast inference, GPU parallelism, and continuous trajectory generation—while drastically improving scalability. H-NTFields improves SR by 3× over NTFields/P-NTFields while remaining competitive in runtime.
In contrast to explicit graph- or grid-search methods such as PRM and FMM, our approach learns a smooth value function that can be queried efficiently once trained.

Fig.~\ref{gibson} illustrates performance in a challenging multi-floor Gibson scene. The example shows a path from the first to the third floor, traversing stairs, doors, and narrow passages. H-NTFields produces a collision-free, smooth trajectory, while other physics-informed methods fail. The generated path avoids the sharp turns often seen in sampling- and grid-based planners.

Together, these results show that H-NTFields bridges the gap between neural-field planners (fast but brittle) and classical planners (robust but slow). It combines the efficiency and parallelism of the former with the scalability and reliability of the latter.

\subsection{Real-world Environments}

We further validate our method in two real-world scenarios that highlight its ability to avoid collisions and generate long-horizon motion plans efficiently.

\textbf{Cabinet manipulation with a UR5e.} In the first scenario, a 7-DOF UR5e arm operates in a tightly packed cabinet (Fig.~\ref{fig:ur5e_arm}). Starting between a screwdriver and a chip cylinder, the arm successfully reaches the second layer without collisions. The collision-free path is generated in 0.07 seconds, demonstrating fast inference in a highly cluttered setting. 

\textbf{Kitchen pick-and-place with a quadruped manipulator.} In the second scenario, a Unitree B1 quadruped equipped with a 7-DOF Z1 arm performs a long-horizon pick-and-place task in a kitchen (Fig.~\ref{fig:arm_real}). The robot begins near a table, grasps a can, and simultaneously coordinates its body and arm to avoid collisions with surrounding chairs before safely placing the can on the target surface. The complete motion plan is generated within 0.15 seconds, showing that our framework can scale to whole-body manipulation in unstructured environments.

\section {Conclusions and Future Work}

This paper presented Hierarchical Neural Time Fields (H-NTFields), a self-supervised framework for learning Eikonal-based value functions for robot motion planning. By combining weak supervision from sparse roadmaps with PDE regularization, our method captures global connectivity while preserving local obstacle-aware geometry, enabling reliable planning in complex multi-room environments. Integrated with sampling-based MPC, H-NTFields provides a reusable continuous cost-to-go representation that supports efficient amortized inference.

While H-NTFields scales reliably when trained per scene, cross-scene generalization remains an open direction for future work. Our current approach does not fully address performance in unseen or structurally diverse environments. A promising direction is to develop more powerful neural encoders or hierarchical scene embeddings that can better transfer across heterogeneous settings.

\bibliographystyle{IEEEtran}
\bibliography{references}

\end{document}